\documentclass[pmlr]{jmlr}

\newif\ifarxiv

\ifarxiv
        \usepackage{fullpage}
        \usepackage{authblk}
        \usepackage{natbib}
        \usepackage{graphicx}
        \pdfoutput=1
\else
        \usepackage{longtable}
        \jmlrvolume{85}
        \jmlryear{2018}
        \jmlrworkshop{Machine Learning for Healthcare}
\fi

\usepackage{hyperref}
\usepackage{color}
\usepackage{amsmath, amssymb}
\usepackage{amsfonts}
\usepackage{mathdots}
\usepackage{booktabs}

\usepackage{custom_formats}

\newcommand{\EX}{\mathbb{E}}
\newcommand{\pdata}{\bx \sim p_\text{data}}
\newcommand{\pz}{\bz \sim p_\bz}

\newif\ifallowcomments

\allowcommentstrue

\ifallowcomments
        \newcommand{\comm}[1]{\textbf{\color{blue} (RM: #1)}}
        \newcommand{\adam}[1]{\textbf{\color{red} (AD: #1)}}
        \newcommand{\ab}[1]{\textbf{\color{cyan} (AB: #1)}}
        \newcommand{\tcyc}[1]{\textbf{\color{red} (TC: #1)}}
\else
        \newcommand{\comm}[1]{}
        \newcommand{\adam}[1]{}
        \newcommand{\ab}[1]{}
        \newcommand{\tcyc}[1]{}
\fi

\begin{document}

\title[Automated Radiation Therapy using Generative Adversarial Networks]{Automated Treatment Planning in Radiation Therapy using Generative Adversarial Networks}

\ifarxiv
\else
        \author{\Name{Rafid Mahmood} \Email{rmahmood@mie.utoronto.ca} \\
                \addr Department of Mechanical and Industrial Engineering \\
                University of Toronto, Toronto, ON, Canada
                \AND
                \Name{Aaron Babier} \Email{ababier@mie.utoronto.ca} \\
                \addr Department of Mechanical and Industrial Engineering \\
                University of Toronto, Toronto, ON, Canada
                \AND
                \Name{Andrea McNiven} \Email{andrea.mcniven@rmp.uhn.ca} \\
                \addr Radiation Medicine Program \\
                Princess Margaret Cancer Centre, Toronto, ON, Canada
                \AND
                \Name{Adam Diamant} \Email{adiamant@schulich.yorku.ca} \\
                \addr Schulich School of Business \\
                York University, Toronto, ON, Canada
                \AND
                \Name{Timothy C. Y. Chan} \Email{tcychan@mie.utoronto.ca} \\
                \addr Department of Mechanical and Industrial Engineering \\
                University of Toronto, Toronto, ON, Canada
        }
\fi

\maketitle

\begin{abstract}
        Knowledge-based planning (KBP) is an automated approach to radiation therapy treatment planning that involves
        predicting desirable treatment plans before they are then corrected to deliverable ones.
        We propose a generative adversarial network (GAN) approach for predicting desirable 3D dose distributions that eschews the previous paradigms of site-specific feature engineering and predicting low-dimensional representations of the plan. Experiments on a dataset of oropharyngeal cancer patients show that our approach significantly outperforms previous methods on several clinical satisfaction criteria and similarity metrics.
\end{abstract}

\section{Introduction}

Radiation therapy (RT) is one of the primary methods for treating cancer and is recommended for over 50\% of all cancer patients~\citep{delaney:2005role}. In RT, a linear accelerator (linac) outputs high-energy x-ray beams from multiple angles around a patient to deliver a prescribed dose of radiation to a tumor while minimizing dose to the healthy tissue.
An RT treatment plan is the result of a complex design process involving multiple medical professionals and several software systems. This includes specialized optimization software that determines the beam characteristics (e.g., aperture shapes for each beam angle, dose delivered from each aperture) required to deliver the final dose distribution. The optimization model takes as input a set of computed tomography (CT) images of the patient, various dosimetric objectives and constraints, and other parameters that guide the optimization process.
The model outputs a treatment plan that is subsequently evaluated by an oncologist. The oncologist usually proposes modifications to the plan, which then requires the treatment planner to re-solve the optimization model using updated parameters. The total process is labor intensive, time-consuming, and costly, as the back-and-forth between the planner and oncologist is often repeated multiple times until the plan is finally approved.

\begin{figure}[t]
        \centering
        \includegraphics[width=0.93\linewidth]{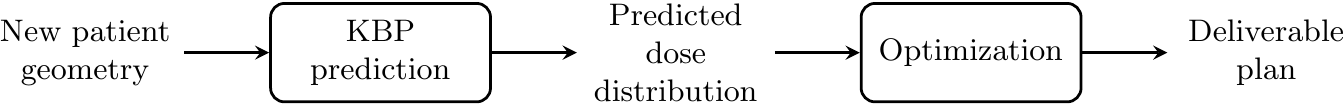}
        \caption{Overview of KBP-driven automated treatment planning pipeline.}
        \label{fig:process}
\end{figure}

The significant manual effort associated with the current treatment planning paradigm, along with the fact that RT plans are generally quite similar for patients with similar geometries, has motivated researchers to investigate how automation can be used in the planning process~\citep{sharpe2014within}. A key enabler of automation is known as knowledge-based planning (KBP), which leverages historically delivered treatments to generate new plans for similar patients. Figure~\ref{fig:process} depicts the two main components of a KBP-driven automated planning system: (i) a machine learning model that uses CT-derived patient geometric features to predict a clinically acceptable three-dimensional dose distribution~\citep{appenzoller:2012predicting,Yang:2013aa,Shiraishi:2015aa,Younge:2018aa}; and (ii) an optimization model that converts the prediction into a ``deliverable'' plan~\citep{McIntosh:2017ab,Wu:2017aa,Babier2018b}. The second step is needed to ensure the treatment plan produced by the machine learning model satisfies the physical delivery constraints imposed by the linac.


A major drawback of most existing KBP prediction methods is their reliance on low-dimensional hand-tailored features derived from patient geometry to predict new dose distributions. In contrast, we propose a new paradigm for generating KBP predictions that automatically learns to predict a 3D dose distribution directly from a CT image. More specifically, we recast the dose prediction problem as an image colorization problem, which we solve using a generative adversarial network (GAN)~\citep{Goodfellow:2014generative}. GANs, which have produced impressive results in other image colorization applications~\citep{isola:2017image,zhu:2017unpaired}, involve a pair of neural networks: a generator that performs a task and a discriminator that evaluates how well the task is performed. In our application, the generator serves as a treatment planner that designs a treatment, while the discriminator plays the role of the oncologist who critiques the generated dose distribution by comparing it to the real treatment plan. Both neural networks train simultaneously on historical data, effectively replicating and aggregating the combined knowledge gained during the iterative manual process used to design clinically acceptable treatments.

In this paper, we develop a novel automated treatment planning pipeline for oropharyngeal cancer that uses a GAN to predict 3D dose distributions. In contrast to previous machine learning methods, our approach does not require the pre-specification of an extensive set of feature variables for prediction. Instead, our model learns what features are important to produce clinically acceptable treatment plans. 
We apply our KBP methodology to a dataset consisting of 26,279 CT images from 217 patients with oropharyngeal cancer that have undergone radiation therapy. Approximately 60\% of these images are used to train the GAN, which is used to predict high quality dose distributions for the remaining out-of-sample patients. These predictions are used as input into an optimization model to produce deliverable plans. We compare our approach to several other techniques, including three feature-based machine learning models and a standard convolutional neural network (CNN). We demonstrate that our approach outperforms all other models in achieving several clinically relevant criteria and in matching the clinical (benchmark) plans.


\paragraph{Technical Significance}
We demonstrate the first use of GANs for generating radiation treatment plans in cancer. We recast KBP prediction as an image colorization problem for which GANs are known to perform well. Moreover, we provide the first full pipeline comparison between different KBP prediction methods by optimizing the predicted dose distribution and comparing the final result to deliverable plans. We find that, in this setting, our GAN approach outperforms all other methods, including the latest in machine learning-based KBP approaches, in meeting clinical criteria.


\paragraph{Clinical Relevance}
Oropharyngeal cancer is one of the most difficult cancers to plan a treatment for, and as a result, generating deliverable treatment plans is particularly time consuming~\citep{Das:2009aa}. Our GAN approach automates the planning approach producing, on average, plans that are superior to clinical ones in several key metrics.
Our site-independent method suggests similar performance for simpler sites, such as prostate and stomach cancers, while showing that high-quality oropharynx treatment plans can be automatically generated.

\section{Related work}

\subsection{Knowledge-based planning}
Many different approaches have been tested for the machine learning component of a KBP-driven automated planning pipeline (cf. Figure~\ref{fig:process}).
Query-based methods
identify previously treated patients who are sufficiently similar to the new patient, and use the historically achieved dose metrics as predictions for the new patient~\citep{Wu:2009aa,Wu:2011aa}. Another common approach uses principal component analysis (PCA), in conjunction with linear regression, to predict dose metrics for new patients~\citep{PCAZhu:2011aa, PCAYuan:2012aa}. However, these well-established techniques only predict two-dimensional dose metrics. Recent research has shown that 3D dose distribution predictions can also be generated using random forest or neural network-based models~\citep{Shiraishi:2016aa,McIntosh:2017aa,nguyen2017dose}.
Nevertheless, for many of these approaches to work effectively, significant effort must be spent in feature engineering, i.e., introducing features specific to the cancer site. Furthermore, some of these approaches compare the predicted dose distributions, rather than deliverable plans post-optimization, to the clinical plans.

For the optimization phase of KBP, there are two main approaches for turning predictions into treatments: dose mimicking~\citep{petersson:2016evaluation} and inverse optimization~\citep{taewoo}. The dose mimicking model minimizes the $L_2$ loss between the predicted dose distribution and one that satisfies all physical constraints. Alternatively, inverse optimization (IO) is a methodology that estimates parameters of an optimization problem from its observed solutions~\citep{ahuja:2001inverse}. In the RT context, IO finds parameters, e.g., objective function weights, that allow a deliverable treatment plan to re-create the predicted dose distribution as closely as possible~\citep{taewoo}. A key advantage of inverse optimization is that it better replicates the trade-offs implicit in clinical treatment plans~\citep{chan2017:trade}.

\subsection{Generative adversarial networks}
GANs are a well-studied class of deep learning algorithms used in \emph{generative} modeling, i.e., in the creation of new data~\citep{Goodfellow:2014generative}. Although initially used to artificially generate 2D images, and later 3D models~\citep{wu:2016learning}, their success has garnered increasing interest for healthcare applications. GANs have been used for medical drug discovery~\citep{kadurin:2017drugan}, generating artificial patient records~\citep{choi:2017gan,esteban:2017gan}, the detection of brain lesions \citep{alex2017generative}, and image augmentation for improved liver lesion classification~\citep{frid:2018gan}.

A GAN consists of two neural networks, a generator and a discriminator, working in tandem. The generator $G(\cdot)$ takes an initial random input $\pz$ and attempts to generate an artificial data sample $\bx = G(\bz)$ (i.e., the 3D dose distribution). The discriminator $D(\cdot)$ is a classifier that takes generated and real data samples, and tries to identify which is which, i.e., $D(\bx) \in [0, 1]$ where $1$ suggests the generated sample is satisfactory. The interaction between the networks can be formalized mathematically as a minimax game. If $\pdata$ is the probability distribution over the real data samples, then the game is defined as
\begin{align*}
        \min_G \max_D \Big\{ V(G, D) = \EX_{\pdata}\left[ \log D(\bx) \right] + \EX_{\pz}\left[ \log (1-D(G(\bz))) \right] \Big\}.
\end{align*}

GANs have been proven effective in style transfer problems, where the generator input $\bz$ is a data sample corresponding to one style (or characteristic) and the output $\bx$ is a mapping to a different style~\citep{isola:2017image,zhu:2017unpaired}. For example, style transfer can be used to transform grayscale images to colored photos~\citep{sangkloy2017scribbler}, in facial recognition for surveillance-based law enforcement~\citep{wang2017back}, and in 3D reconstruction of damaged artifacts~\citep{hermoza:20173d}. Here, the generator $G(\bz)$ learns the mapping between styles that generates samples resembling the ground truth. Since key structures in the output may be entangled with noise from the generator, the desired output is often achieved by modifying the original minimax game with a penalty term on large deviations between the real and generated samples:
\begin{align} \label{styleTransferLoss}
        \min_G \max_D \Big\{V(G,D) + \lambda \EX_{\pdata, \pz} \left[ \| \bx - G(\bz)  \|_1 \right] \Big\},
\end{align}
where $\lambda$ is a regularizer that balances the trade-off between learning style and the real data.

\section{Methods}

\begin{figure}[t]
        \centering
        \includegraphics[width=0.94\linewidth]{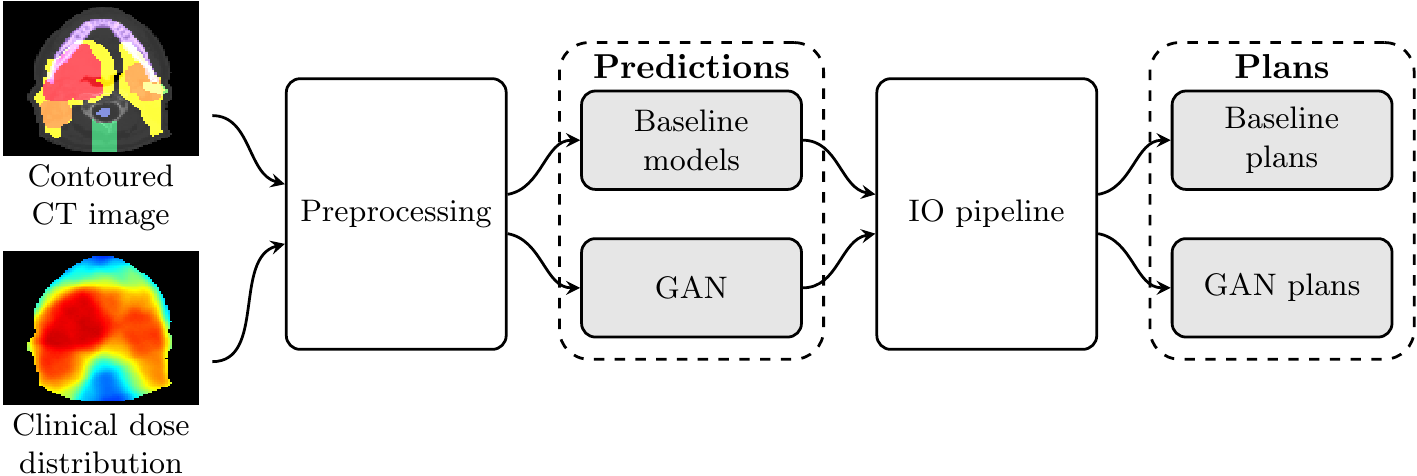}
        \caption{An schematic of our KBP-based automated planning pipeline.}
        \label{fig:methods}
\end{figure}

We used contoured CT images and clinically acceptable dose distributions from the treatment plans of past oropharyngeal cancer patients to train a style transfer GAN. We then passed out-of-sample predicted dose distributions through an IO pipeline~\citep{Babier2018b} to generate the final treatment plans. For baseline comparisons, we also implemented several methods from the literature using the complete pipeline. Figure~\ref{fig:methods} shows a high-level overview of this automated planning pipeline.

\subsection{Data}

We obtained treatment plans from 217 oropharyngeal cancer patients treated at a single institution with 6 MV, step-and-shoot, intensity-modulated radiation therapy machine. All plans were for a prescription of 70 Gy, 63 Gy, and 56 Gy in 35 fractions to the gross disease, intermediate risk, and elective target volumes, respectively.

For each patient, we identified a set of targets and healthy organs-at-risk (OARs). Targets were denoted as planning target volumes (PTVs) along with the oncologist-prescribed dose (e.g., PTV70 corresponds the target with the highest dose prescription). OARs included the brainstem, spinal cord, right and left parotids, larynx, esophagus, and mandible. Every voxel (a 3D pixel of size $4$ mm $\times$ $4$ mm $\times$ $2$ mm) of a CT image was classified by their clinically drawn contours. All voxels were assigned a structure-specific color, and in cases where the voxel was classified as both target and OAR, we reverted to target. 
All unclassified tissue was left as the original CT image grayscale.

\subsection{GAN model}
We first divided each 3D CT image into 2D slices of $128\times 128$ pixels. The generator used a single CT image slice to predict the dose distribution along that same plane without considering the vertical relationship between different slices. This process was repeated for every slice until a full 3D dose distribution was produced. Our training set consisted of all 2D slices from the 3D CT images for 130 patients, totaling 15,657 images. The CT images from the remaining 87 patients were used for out-of-sample evaluation.

Our GAN learning model was built on the \texttt{pix2pix} style transfer architecture of~\citet{isola:2017image}. We used a U-net generator that passed a 2D contoured CT image slice through consecutive convolution layers, a bottleneck layer, and then through several deconvolution layers. The U-net also employed skip connections, i.e., the output of each convolution layer was concatenated to the input of a corresponding deconvolution layer. This allowed the generator to easily pass ``high dimensional'' information (e.g., structural outlines) between the inputted CT image slice and the outputted dose slice. The discriminator passed a 2D slice of the dose distribution along several consecutive convolution layers, outputting a single scalar value. In the training phase, the discriminator received one real and one generated dose distribution before backpropagation. We disconnected the discriminator after training, at which point the generator only received a contoured CT slice. We refer the reader to Appendix A for additional details regarding the network architectures.

We used the loss function given by \eqref{styleTransferLoss} with $\lambda=90$, and trained using Adam~\citep{kingma:2014adam}, with learning rate $0.0002$ and $\beta_1=0.5$ and $\beta_2=0.999$ for $25$ epochs. We used the default Adam settings from~\citet{isola:2017image}, as they were proven to be good for a variety of different style transfer problems. While we swept through various values for $\lambda$ and the number of epochs, we found these default settings to be sufficient, with minimal subsequent improvement. We found it useful to stop training when the loss functions were roughly equal; if the loss from the $l_1$ penalty fell too low, the GAN began to simply memorize the dataset. The code for all experiments, along with the parameter settings is provided at~\url{http://github.com/rafidrm/gancer}.

\subsection{Plan generation}

Predicted dose distributions were inputted into an IO pipeline to generate optimized plans. The IO model determined the weights of a parametric ``forward'' optimization model given a predicted dose distribution. The objective of the forward model was to minimize the sum of 65 objective functions: seven per OAR and three per target. Terms for the OARs included the mean dose, max dose, and the percentile (0.25, 0.50, 0.75, 0.90, and 0.975) above the maximum predicted dose to the OAR. Similarly, terms for the target included the maximum dose, average dose below prescription, and average dose above prescription. The complexity of the KBP-generated treatment plan was constrained to match the clinical treatment~\citep{Craft:2007aa} where complexity represents a (convex) surrogate measure for the physical deliverability of a plan. We note that in reality, there are additional constraints in the IO pipeline that we omit for tractability. Thus, our notion of a deliverable plan does not include all physical constraints.
Physical parameters for the optimization model were derived from \texttt{A\ Computational\ Environment\ for\ Radiotherapy\ Research} \citep{CERR}. To replicate the clinical plans, all KBP-generated plans were delivered from nine equidistant coplanar beams at angles 0$^\circ$, 40$^\circ$, \ldots, 320$^\circ$. 
We used \texttt{Gurobi 7.5} to solve the inverse and forward optimization problems associated with the IO pipeline. Additional details of the IO model can be found in~\citet{Babier2018a}.

\subsection{Baseline approaches}

We compared our GAN approach to generating predicted dose distributions with several state-of-the-art techniques. We briefly describe the baseline approaches here.
\begin{itemize}
        \item \textbf{Bagging query (BQ):} A look-up method identifies patients with similar geometries who have undergone radiation therapy and  outputs their doses as predictions. This approach predicts dose volume histograms (DVHs), i.e., 2D summaries of the 3D dose delivered to specific targets and OARs (e.g., \citet{Wu:2009aa,Babier2018a}).

        \item \textbf{Generalized PCA (gPCA):} A method combining PCA with linear regression using patient geometry features. Similar to BQ, this method also predicts DVHs (e.g., \citet{PCAYuan:2012aa,Babier2018a}).

        \item \textbf{Random forest (RF):} Predicts dose to each voxel (3D dose prediction) using ten customized features based on patient geometry (inspired by \citet{McIntosh:2017aa}). Additional details can be found in Appendix B.

        \item \textbf{U-net (CNN):} Predicts dose to each voxel in 2D slices from a CT image using a U-net convolution neural network architecture (e.g., \citet{nguyen2017dose}).
\end{itemize}
All baseline predictions were fed into the same IO pipeline as the GAN approach to ensure a fair comparison between deliverable plans.

\section{Results}
\subsection{Sample generated dose distributions}

\begin{figure}[htb]
        \centering
        \includegraphics[width=0.94\linewidth]{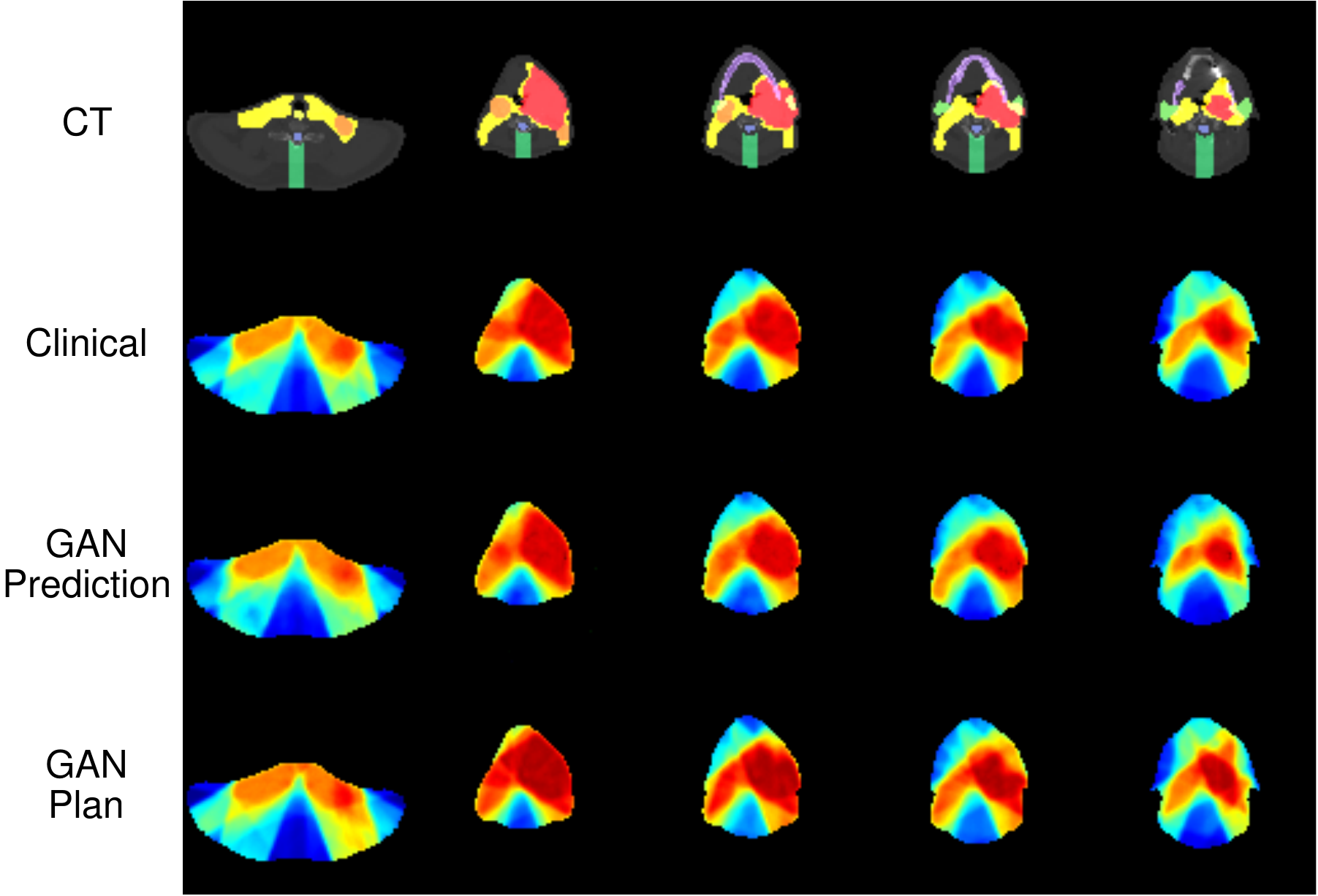}
        \caption{Sample of slices from a test patient. From top to bottom: contoured CT image (generator input), clinical plan (ground truth), GAN prediction, and GAN plan (post optimization).}
        \label{manifold}
\end{figure}

We observed that the style transfer function mapping the CT image to the predicted dose distribution appeared easy to learn. This is because the GAN generated dose distributions had the hallmarks of a deliverable plan, like the sharp dose gradients that are generated by individual beams. However, there were subtle deliverability characteristics that the GAN could not always identify. The optimization step enforced these physical deliverability constraints to correct for these idiosyncracies. This result can be observed in Figure~\ref{manifold}, where five sample slices of a clinical, predicted, and optimized plan are presented.

\subsection{Clinical criteria satisfaction}

We measured plan quality by evaluating how frequently they satisfied the standard clinical criteria for oropharyngeal cancer treatment plans; see Table~\ref{tab:goals}. Clinicians commonly use criteria satisfaction as a metric to evaluate plan quality and approve a treatment plan after it satisfies a sufficient number of the criteria. Thus, each criterion (one per OAR and target) was measured on a pass-fail basis depending on whether the mean dose $\mathcal{D}_{mean}$, maximum dose $\mathcal{D}_{max}$, or the dose to $99\%$ of the volume of that structure $\mathcal{D}_{99}$, was above or below a given threshold. To facilitate the comparisons, we scaled the GAN and baseline treatment plans so that their PTV $\mathcal{D}_{99}$ was equal to the PTV $\mathcal{D}_{99}$ of the corresponding clinical plan.

\begin{table}[htb]
        \centering
        \begin{tabular}{c c} \toprule
                Structure & Criteria \\ \midrule
                Brainstem & $\mathcal{D}_{max} \le$ 54 Gy \\
                Spinal Cord & $\mathcal{D}_{max} \le$ 48 Gy \\
                Right Parotid & $\mathcal{D}_{mean}\le$ 26 Gy \\
                Left Parotid & $\mathcal{D}_{mean}\le$ 26 Gy \\
                Larynx & $\mathcal{D}_{mean} \le$ 45 Gy \\
                Esophagus & $\mathcal{D}_{mean} \le$ 45 Gy \\
                Mandible & $\mathcal{D}_{max}\le$ 73.5 Gy \\
                PTV56 & $\mathcal{D}_{99}\ \ge$ 53.2 Gy \\
                PTV63 & $\mathcal{D}_{99}\ \ge$ 59.9 Gy \\
                PTV70 & $\mathcal{D}_{99}\ \ge$ 66.5 Gy \\\bottomrule
        \end{tabular}
        \caption{Clinical criteria used to evaluate all plans. $\mathcal{D}_{mean}$ refers to the mean dose, $\mathcal{D}_{max}$ the maximum dose, and $\mathcal{D}_{99}$ dose to $99\%$ of the structure.}
        \label{tab:goals}
\end{table}

Table~\ref{ClinSat} presents the percentage of the GAN and baseline treatment plans that satisfied the clinical criteria. We note that clinically acceptable plans typically cannot satisfy all criteria simultaneously because of the proximity of the targets to the OARs and the complexity of the head-and-neck site in general. We observed that the BQ and gPCA plans tended to satisfy PTV criteria more frequently, which suggested that they may recommend delivering a higher dose to the target relative to the clinical plan. However, they failed to achieve mean and maximum dose criteria to the OARs (note: there are more than triple the number of OAR criteria as PTV criteria once all plans are normalized to $\mathcal{D}_{99}$ of the PTV70). On the other hand, the RF plans appeared to satisfy fewer clinical criteria associated with the target as compared to the clinical plans. The CNN plans achieved the closest level of performance to the clinical plans. However, the GAN plans had the best overall performance among all approaches. They offered a balanced trade-off between the OARs and targets, and even outperformed the clinical plans on clinical criteria satisfaction.


\begin{table}[htb]
        \centering
        \begin{tabular}{lcccccc} \toprule
                &  BQ & gPCA & RF & CNN & GAN & Clinical \\ \midrule
                OAR criteria       &   61.6\%  & 65.8\%& 71.5\% & 72.5\%&72.8\% & 72.0\%\\
                PTV criteria  &  83.5\%  & 85.7\% & 68.0\% &76.3\% & 81.3\% & 76.8\%\\
                All criteria & 67.6\%  & 71.2\% & 70.7\% & 73.6\% & 75.2\% & 73.3\%\\ \bottomrule
        \end{tabular}
        \caption{Frequency of clinical criteria satisfaction.}
        \label{ClinSat}
\end{table}

\begin{figure}[htb]
        \centering
        \subfigure[BQ $-$ clinical]{%
        \includegraphics[width=0.31\linewidth]{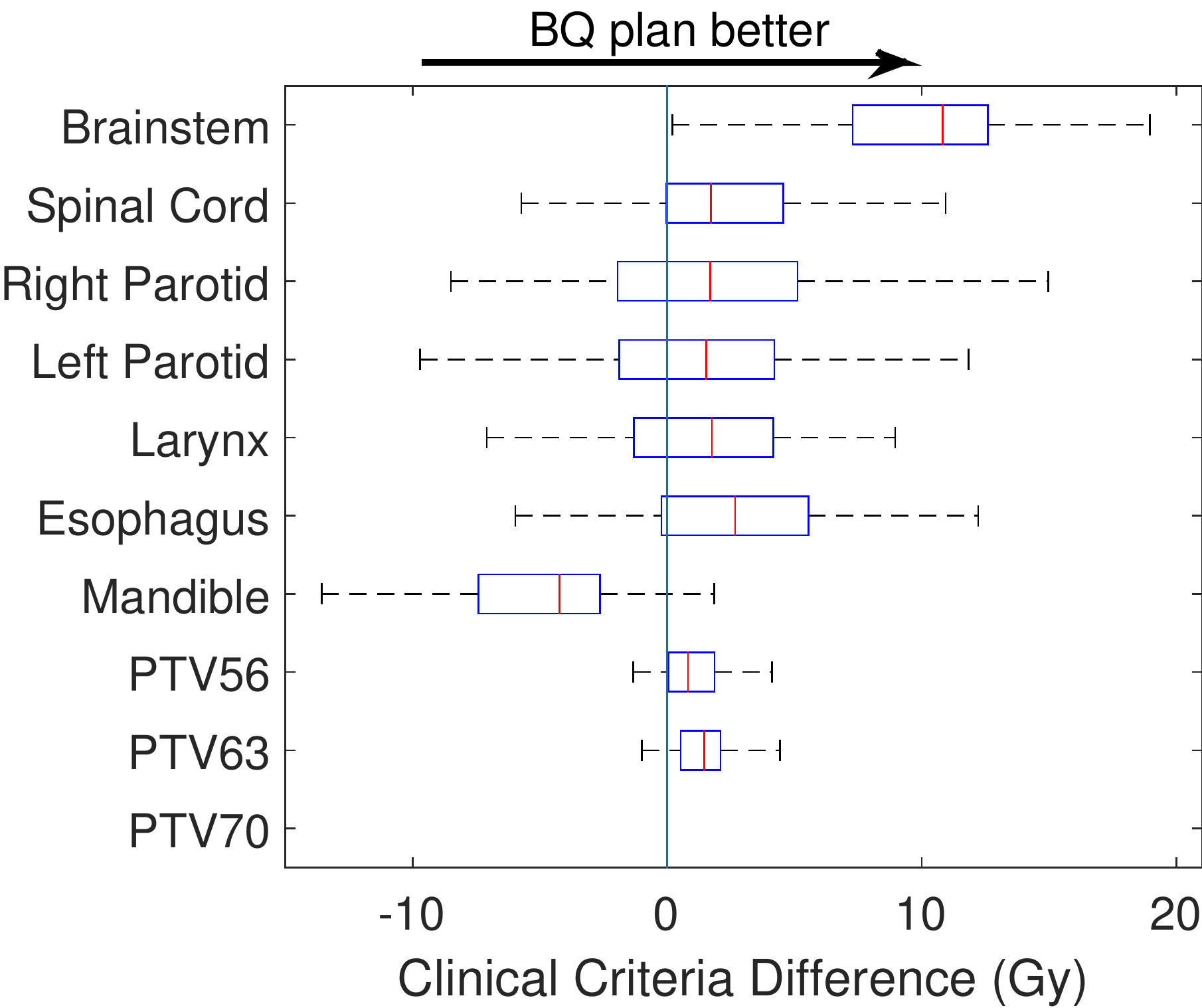} }
        \subfigure[gPCA $-$ clinical]{%
        \includegraphics[width=0.31\linewidth]{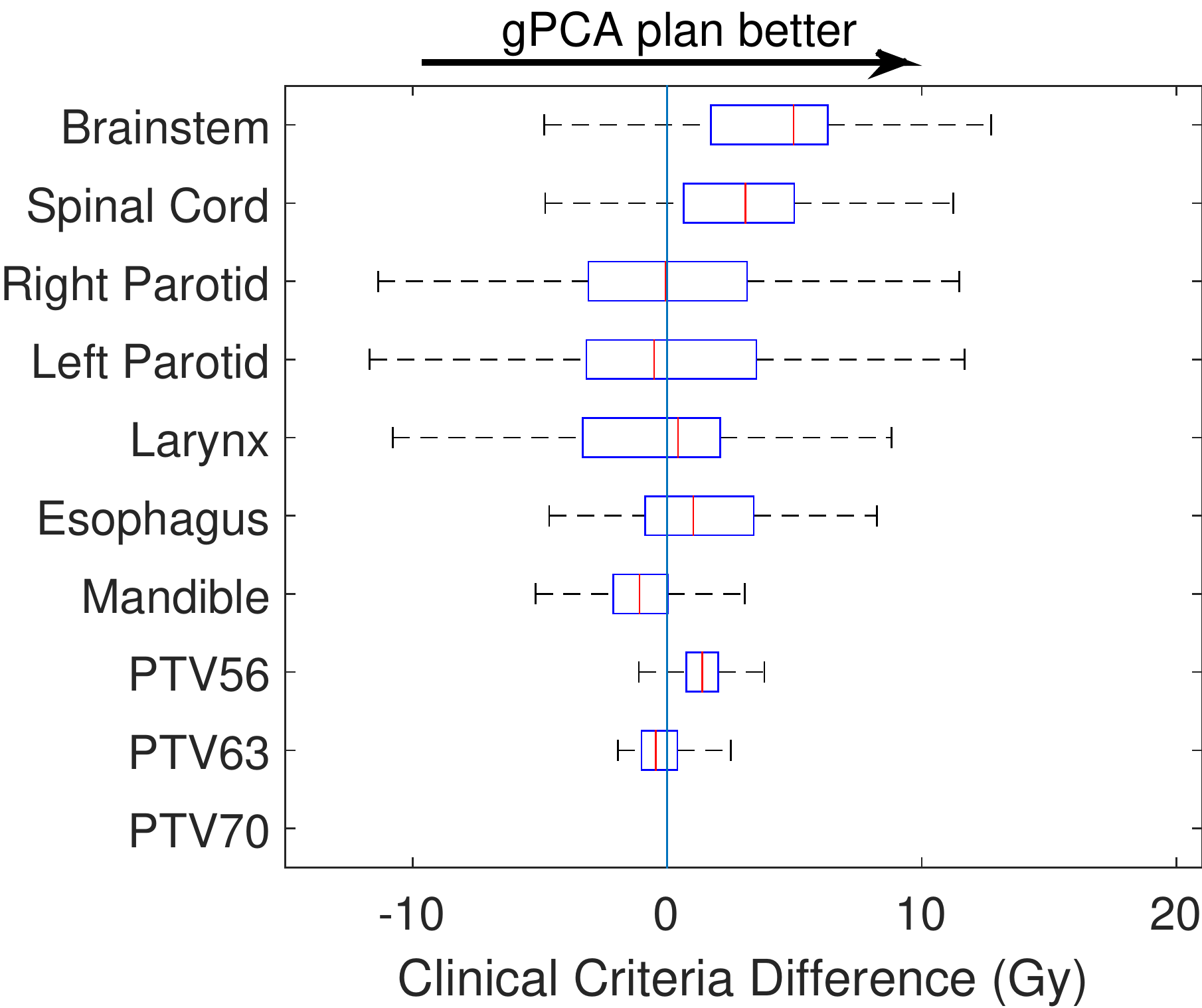} }
        \subfigure[RF $-$ clinical]{%
        \includegraphics[width=0.31\linewidth]{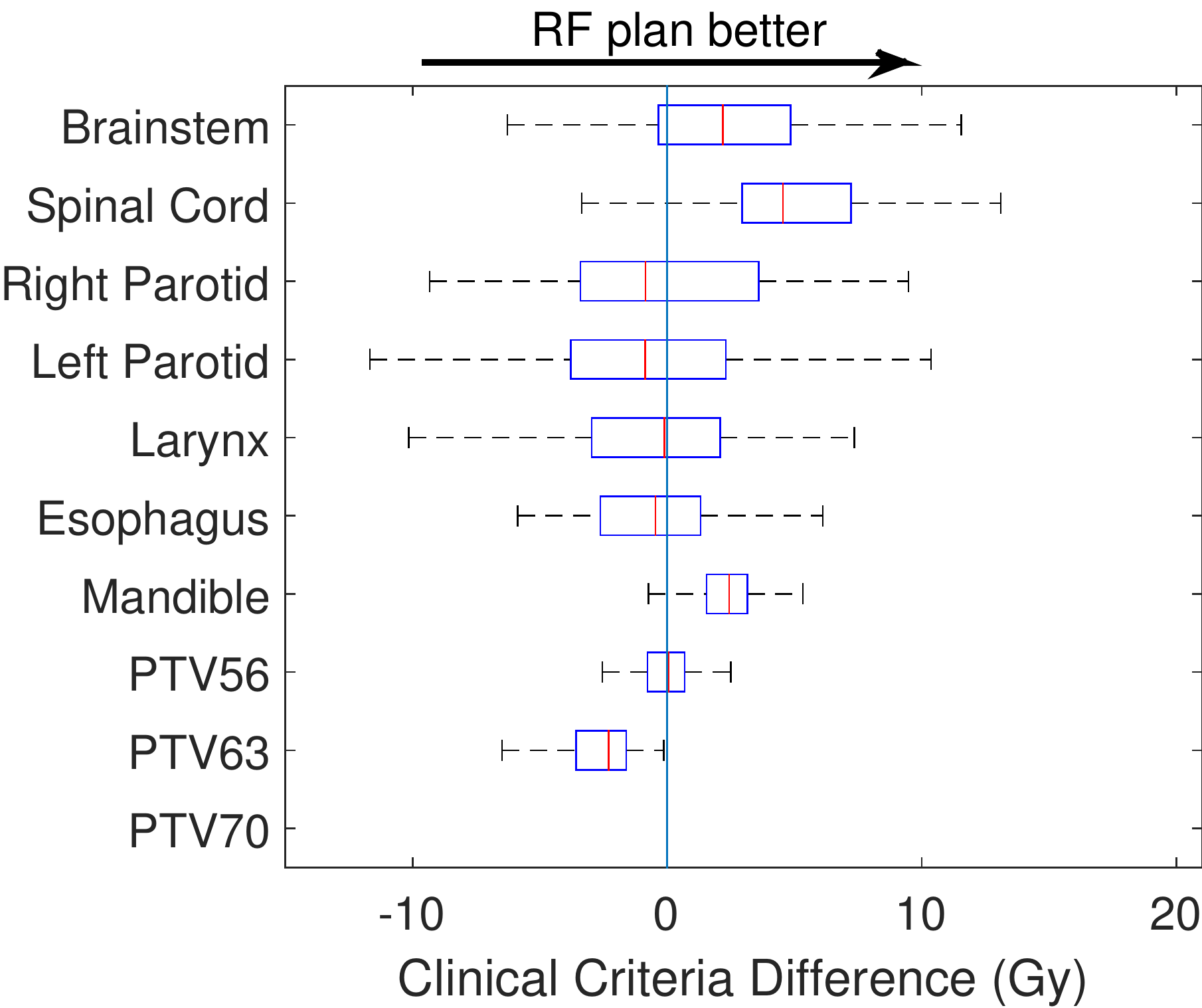} }
        \subfigure[CNN $-$ clinical]{%
        \includegraphics[width=0.31\linewidth]{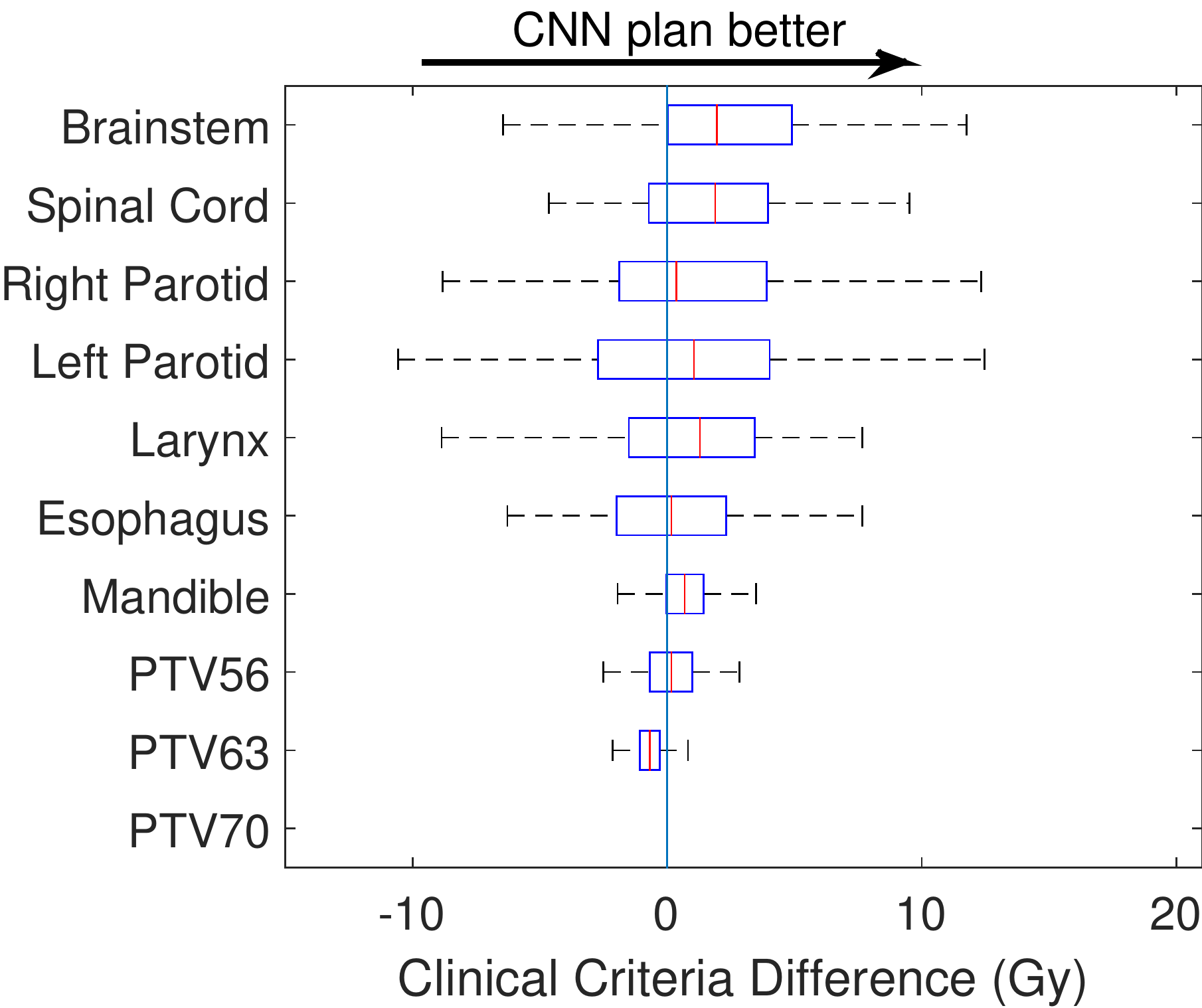} }
        \subfigure[GAN $-$ clinical]{%
        \includegraphics[width=0.31\linewidth]{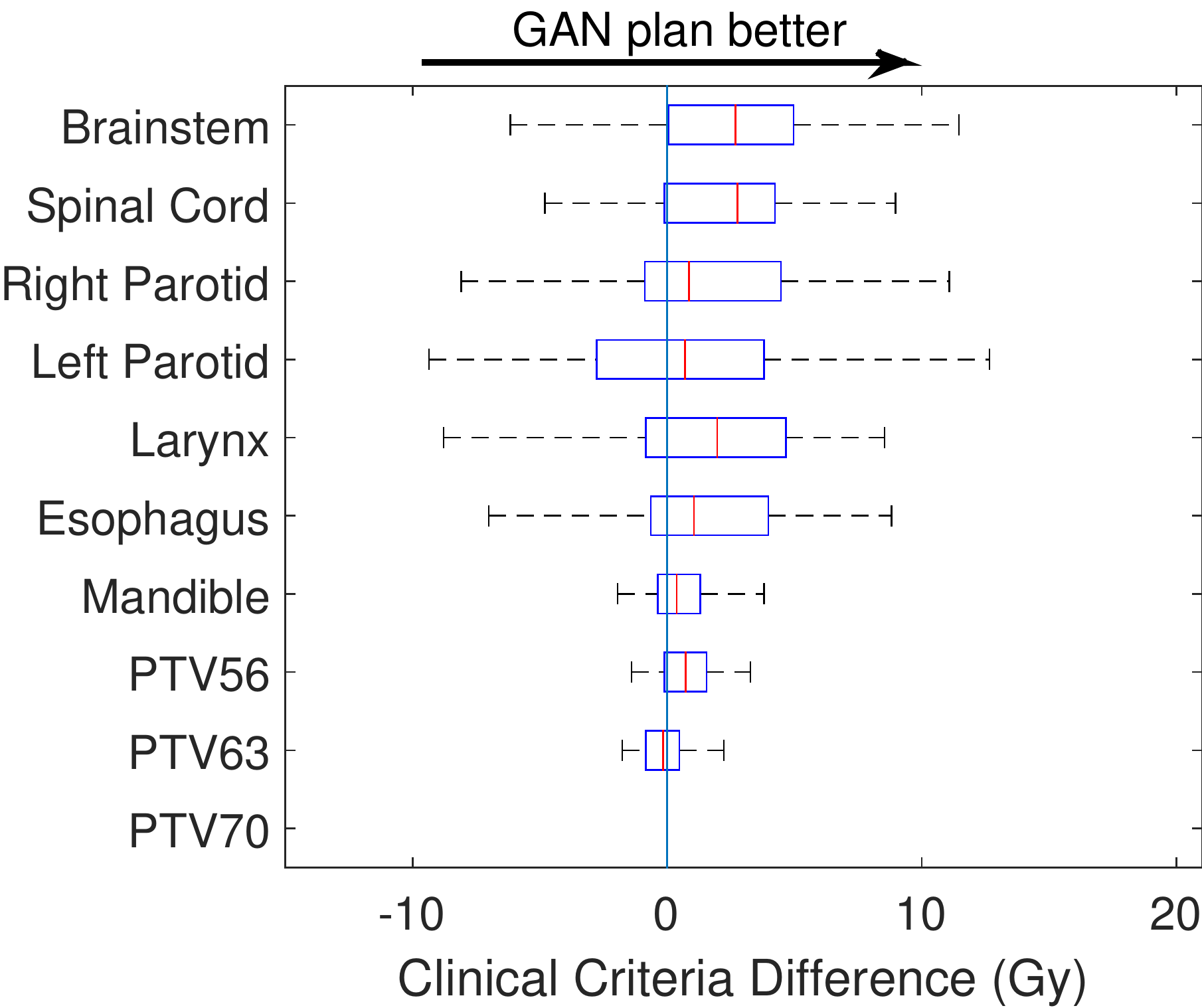} }
        \subfigure[GAN $-$ CNN]{%
        \includegraphics[width=0.31\linewidth]{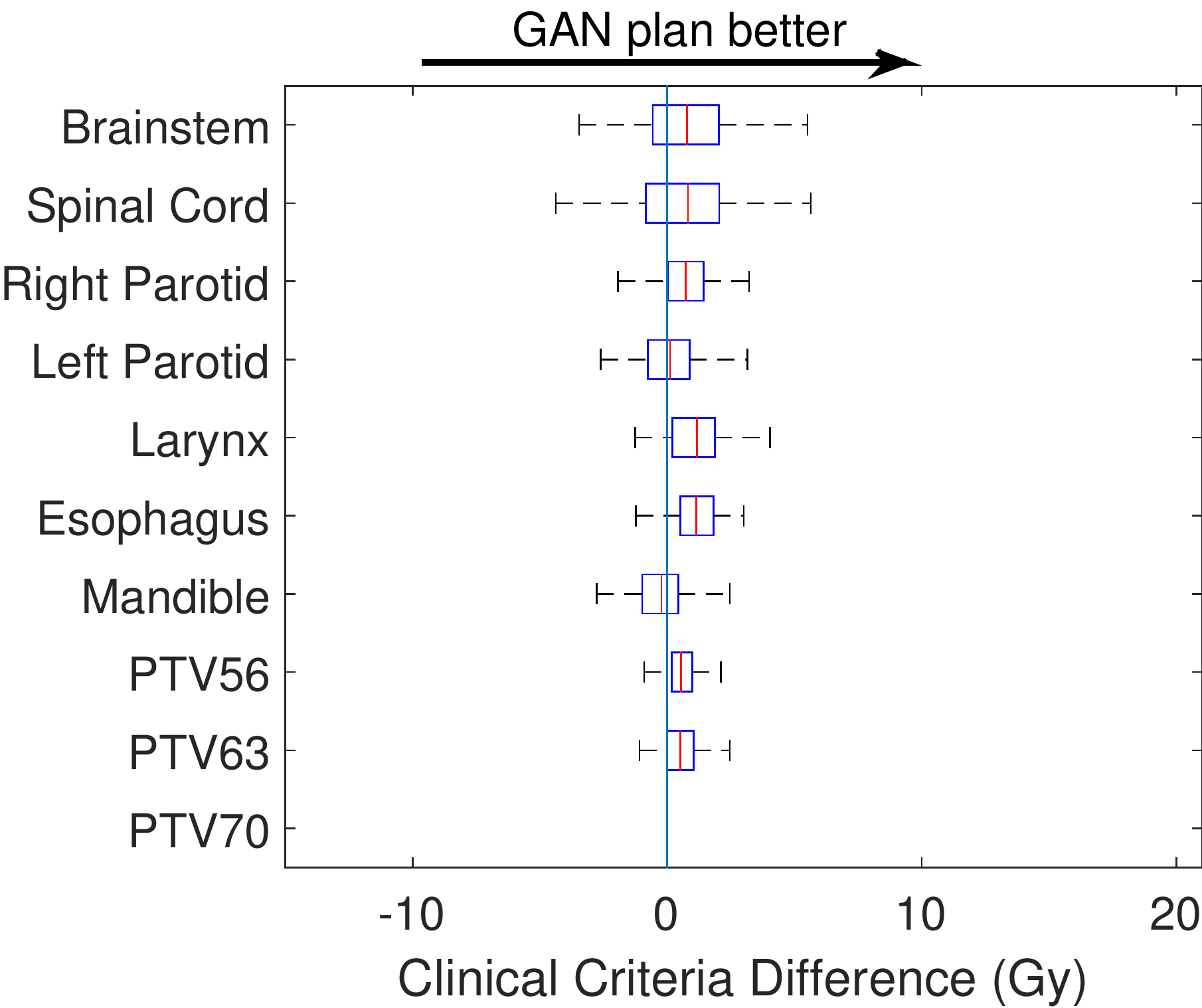} }
        \caption{Head-to-head comparisons: (a)--(e) the plans from each KBP-generated model versus their clinical counterparts where positive difference implies the KBP-generated plans were better; (f) the plans from the GAN versus the CNN. Upper and lower boundaries of each box represent the 75th and 25th percentiles respectively, and the vertical line in the box depicts the median. Whiskers extend to 1.5 times the interquartile range. The line across each plot provides a reference for zero difference.}
        \label{clinPerf}
\end{figure}

The previous results focused on pass-fail performance with respect to the clinical criteria. We also examined the magnitude of passing or failing via head-to-head comparisons of the GAN/baseline plans to the clinical plans, and between the GAN and CNN plans (see Figure~\ref{clinPerf}). The x-axis in each figure is the difference in Gray (Gy) between the KBP and the clinical plans (KBP minus clinical) for the criterion on the corresponding y-axis. We found that for each criterion, the majority of GAN plans outperformed their clinical counterparts by several Gy (Figure~\ref{clinPerf}(e)). This is a significant result given that the clinical plans were heavily optimized and delivered to actual patients. The BQ, gPCA, and RF plans displayed substantial variability in performance when compared to the clinical plan. Consistent with Table~\ref{ClinSat}, performance of the CNN plans were closest to the GAN plans although, as shown in Figure~\ref{clinPerf}(f), the GAN plans maintained a small, yet consistent, advantage.

Finally, we compared the KBP plans against the clinical plans using the gamma passing rate (GPR) metric. GPR measures the similarity between two dose distributions on a voxel-by-voxel basis, computing for each voxel, a pass-fail test. We considered the standard choice of GPR, i.e., a 3\%/3 mm tolerance~\citep{low1998technique}, which roughly means a voxel in the evaluated dose distribution (KBP) ``passes'' if there is at least one voxel in the reference dose distribution (clinical) within $3$ mm that receives a dose that is within $\pm$3\% of the reference dose. Table~\ref{GPR} summarizes the average GPR achieved over all KBP-generated plans. A score of $1.0$ means that every voxel has passed the criteria; in other words, the two dose distributions were considered identical (within the tolerance). Overall, we observed that the GAN plans generated dose distributions that most closely resembled the clinical dose distributions, followed by the CNN, and then the gPCA plans. Notably, the GAN dose distributions best resembled the clinical dose distribution around the target, which is of primary importance. The GAN plans performed less well on the OARs, but this result was expected given the results from Table~\ref{ClinSat}, which indicated that the GAN plans achieved more OAR clinical criteria than the clinical plan (i.e., the GAN was able to deliver a lower dose to the OARs as compared to the clinical dose distribution).

\begin{table}[htb]
        \centering
        \begin{tabular}{lcccccc} \toprule
                &  BQ & gPCA & RF & CNN & GAN \\ \midrule
                All OARs   &   0.548 & 0.584& 0.535 & 0.566& 0.549 \\
                All PTVs    &  0.533  & 0.728 & 0.503 & 0.741 & 0.761 \\
                All Structures & 0.536  & 0.669 & 0.518 & 0.670 & 0.675 \\ \bottomrule
        \end{tabular}
        \caption{Average GPR for each population of KBP plans compared to clinical plans. }
        \label{GPR}
\end{table}

\section{Discussion and Future Work}

In this paper, we proposed the first GAN-based KBP method to generate radiation therapy treatment plans. We trained our complete pipeline on 130 patients, tested on 87 out-of-sample patients diagnosed with oropharyngeal cancer, and compared our technique with several state-of-the-art planning methods including a query-based approach, a PCA-based method, a random forest, and a CNN. All methods were evaluated on standard clinical criteria for plan evaluation (i.e., OARs sparing and target coverage), showing that the GAN plans outperformed all baseline KBP methods. We also demonstrated that the GAN plans outperformed the clinical plans by satisfying additional criteria on OAR dose sparing and target dose coverage. Finally, we used the gamma passing rate, a standard metric in the radiation therapy literature, to evaluate the similarity of the full 3D dose distribution between the KBP and clinical plans demonstrating that the GAN plans were the most similar to clinical plans on average. Note that the performance of automated planning methods should be measured based on their ability to re-create clinical quality plans with minimal manual effort. Of course, if the auto-generated plans manage to improve upon the clinical plans, that would be even better.

Our approach eschews the classical paradigm of predicting low-dimensional representations, or engineering features, by training a generic neural network to learn desirable dose distributions.
Specifically, the GAN recasts KBP prediction as an image colorization problem. Moreover, the GAN is trained by mimicking the iterative process between the treatment planner and oncologist; the generator network acts as the treatment planner by designing dose distributions while the discriminator acts as the oncologist by determining whether the plans are good or bad. The implication is that selecting the appropriate neural network architecture may be sufficient when creating an automated KBP pipeline that generates deliverable plans. Further, our approach does not add site-specific feature variables which suggests that the good performance we observe may not be limited to patients with oropharyngeal cancer. Finally, since the GAN plans improve upon the clinical plans, it may be useful to analyze the results to generate useful insights for practitioners.

We envision two interesting directions for future work. First, we plan to explore how GANs can develop treatment plans for different cancer sites. By adding site labels, we expect that a GAN can learn from the augmented training set of different cancer sites to better develop plans for specific sites. Second, we hope to automate the preprocessing stage by using uncontoured CT images. As neural networks show increasing promise for automated image segmentation (i.e., tumor and healthy organ identification), we hope to leverage this work to improve our treatment plan prediction model. 

\acks{This study was approved by the institutional research ethics board. Support for this research was provided by the Natural Sciences and Engineering Research Council of Canada.}

\bibliography{refs}

\appendix
\section*{Appendix A. Network architecture}

The general network architecture was adapted from~\citet{isola:2017image}.   Contoured CT slices were used as input to the generator as 3-channel, $128 \times 128$ images. We used a U-net architecture, where the generator was comprised of an encoder and a decoder stage. We used $4 \times 4$ 2D convolutions with stride $2$ and padding $1$. Each convolution layer was followed by a leaky ReLU and batch normalization. Deconvolution layers were followed by $50\%$ dropout, ReLU, and batch normalization.

The encoder consisted of four downsampling layers. The first generated 64 channels, and each subsequent layer downsampled by a factor of $2$. This was followed by $2$ bottleneck layers, before the data was then passed through $4$ upsampling layers. The output of each downsample layer was concatenated to the input of the corresponding upsample layer. The final output was a 3-channel, $128 \times 128$ slice.

The decoder consisted of five convolution layers, where the first four each downsample the output by $2$. The fifth, and last layer, mapped to a scalar output. Once again, we applied batch normalization and leaky ReLU after the first four layers. The final layer was passed through sigmoid activation.

\section*{Appendix B: Random forest architecture}

\begin{table}[htb]
        \centering
        \begin{tabular}{c l} \toprule
                Feature & Description \\ \midrule
                Structure & Structure that the voxel is classified as\\
                $y$-coordinate & Voxel's positions on the $y$-axis in a slice\\
                $z$-coordinate & Plane of voxel's slice\\
                Distance to larynx & Shortest path between voxel and the surface of the larynx \\
                Distance to esophagus & Shortest path between voxel and the surface of the esophagus\\
                Distance to limPostNeck & Shortest path between voxel the surface of  the limPostNeck\\
                Distance to PTV56 & Shortest path between voxel and the surface of the PTV56\\
                Distance to PTV63 & Shortest path between voxel and the the surface of  PTV63\\
                Distance to PTV70 & Shortest path between voxel and the the surface of PTV70\\
                Influence & Sum of influence matrix elements for the voxel\\\bottomrule
        \end{tabular}
        \label{rfFeatures}
        \caption{The ten features used in the RF to predict the dose for any voxel.}
\end{table}

The random forest used ten custom features outlined in Table~\ref{rfFeatures} to predict the dose delivered to each voxel in the patient. The RF was trained with ten trees, and default settings with the \texttt{randomForestRegressor} from \texttt{scikit-learn}.




\end{document}